%% file: eamt23.tex
\documentclass[11pt]{article}
\usepackage{eamt23}
\usepackage{times}
\usepackage{url}
\usepackage{latexsym}
\usepackage[small,bf]{caption} 
\usepackage{todonotes}
\usepackage{booktabs}
\usepackage{multirow}
\usepackage{xcolor,colortbl} 

\definecolor{green}{rgb}{0.49, 0.99, 0.0}
\definecolor{red}{rgb}{1.0, 0.0, 0.0}
\definecolor{orange}{rgb}{1.0, 0.5, 0.0}

\newcommand{\debertavt}{\textsc{deberta-v3}}
\newcommand{\xlmr}{\textsc{xlm-r}}
\newcommand{\mbart}{\textsc{m-bart}}
\newcommand{\mbert}{\textsc{m-bert}}
\newcommand{\mdeberta}{\textsc{m-deberta}}
\newcommand{\centdash}{\multicolumn{1}{c}{---}}
\newcommand{\deb}{\textsc{deb}}
\newcommand{\mdeb}{\textsc{m-deb}}
\newcommand{\expnot}[2]{$#1\mathrm{e}^{#2}$}
\newcommand{\stddev}[1]{\small$\pm#1$}

\usepackage{soul}

\title{Automatic Discrimination of Human and Neural\\Machine Translation in Multilingual Scenarios}

\author{Malina Chichirau\\
 Bernoulli Institute\\
  University of Groningen\\
  {\small \tt m.chichirau@student.rug.nl}  \And
  Rik van Noord\\
  CLCG\\
  University of Groningen\\
  {\small \tt rikvannoord@gmail.com}  \And
  Antonio Toral\\
  CLCG\\
  University of Groningen\\
 {\small \tt a.toral.ruiz@rug.nl}}

\date{}

\begin{document}
\maketitle

\begin{abstract}
We 
tackle the task of automatically discriminating between human and machine translations. As opposed to most previous work, we perform experiments in a multilingual setting, considering multiple languages and multilingual pretrained language models. 
We show that a classifier trained on parallel data with a single source language (in our case German--English) can still 
perform well on 
English translations that come from different source languages, even when the machine translations were
produced by other systems than the one it was trained on. 
Additionally, we demonstrate
that incorporating the source text in the input of a multilingual classifier improves (i) its accuracy and (ii) its robustness on cross-system evaluation, compared to a monolingual classifier. Furthermore, we find that using training data from multiple source languages (German, Russian, and Chinese) tends to improve the accuracy of both monolingual and multilingual classifiers.
Finally, we show that bilingual classifiers and classifiers trained on multiple source languages benefit from being trained on longer text sequences, rather than on sentences.

\end{abstract}

\section{Introduction}

 In many NLP tasks
 one may
 want to filter 
 out
 machine translations (MT), but keep human translations (HT). Consider, for example, the construction of parallel corpora used for training
 MT
 systems: filtering 
 out
 MT output is getting progressively harder, given the ever-increasing quality of neural MT (NMT) systems. Moreover, the existence of such high-quality NMT systems might aggravate the problem, as people are getting more likely to employ them when creating texts. In addition, it is also hard to get \emph{fair} training data to build a classifier that can distinguish between these two types of translations, since publicly-available parallel corpora with human translations were likely used in the training of well-known publicly available MT systems (such as Google Translate or DeepL). Therefore, we believe that making the most of the scarcely available (multilingual) training data
 is a crucial research direction.

Previous work aiming at discriminating between HT and NMT operates mostly in a monolingual setting~\cite{fu-nederhof-2021-automatic,van-der-werff-etal-2022-automatic}, 
To our knowledge, the only exception is~\newcite{bhardwaj-etal-2020-human}, who targeted English--French, and fine-tuned not only French LMs (monolingual target-only setting), but also multilingual LMs, so that the classifier had also access to the source text. However, this work used an in-house data set, therefore limiting reproducibility and practical usefulness.
There is also older work that tackled
statistical MT
(SMT) vs HT classification  \cite{arase2013machine,aharoni2014automatic,li2015machine}. 
Nevertheless, since both the MT paradigm (SMT) and the classifiers used are not state-of-the-art anymore, less recent studies are of limited relevance today.

Compared to previous work, this paper explores the classification of HT vs NMT in the multilingual scenario in more depth, considering several languages and multilingual LMs.
We demonstrate that classifiers trained on parallel data with a single source language still work well when applied to translations from other source languages (\textbf{Experiment 1}). We show improved performance for fine-tuning multilingual LMs by incorporating the source text (\textbf{Experiment 2}), which also diminishes the gap between training and testing on different MT systems (\textbf{Experiment 3}). Moreover, we improve performance when training on
additional training data from \textit{different} source languages (\textbf{Experiment 4}) and full documents instead of 
isolated
sentences (\textbf{Experiment 5}).

\section{Method}

\subsection{Data} To get the source texts and human translation part of the data set, we use the data sets provided across the WMT news shared tasks of the past years.\footnote{For example,~\url{https://www.statmt.org/wmt19/translation-task.html}}
As explained in the previous section, we only use the WMT test sets, to (reasonably) ensure that the popular MT systems we will be using (Google Translate and DeepL) did not use this as training data. Note that if any of the MT systems had used this data for training the task would actually be \emph{harder}, since their translations for the data would be expected to resemble more human translations than if this data had not been used for training.
An alternative would be to use in-house datasets, like~\newcite{bhardwaj-etal-2020-human}, but that also comes with an important drawback, namely limited reproducibility.

We run experiments across 7 language pairs (German, Russian, Chinese, Finnish, Gujarati, Kazakh, and Lithuanian to English) and use only the source texts that were originally written in the source language, following the findings by~\newcite{zhang-toral-2019-effect}. The data of WMT19 functions as the test set for all languages, while WMT18 is the development set (only used for German, Russian and Chinese, since the other languages are tested in a zero-shot fashion). 
Detailed data splits are shown in Table~\ref{tab:data}.

\begin{table}[!htb]
    \centering
    \setlength{\tabcolsep}{5pt}
    \resizebox{0.975\columnwidth}{!}{
    \begin{tabular}{lrrr}
         \toprule
         & \bf Train & \bf Dev & \bf Test \\
         \midrule
          \bf Sentence-level & & & \\
          \hspace{10pt} German (WMT08-19) & 8,242 & 1,498 & 2,000 \\
          \hspace{10pt} Russian (WMT15-19) & 4,136 & 1,386 & 1,805 \\
          \hspace{10pt} Chinese (WMT17-19) & 878 & 2,260 & 1,838 \\
          \hspace{10pt} Finnish (WMT19) & \centdash & \centdash & 1,996 \\
          \hspace{10pt} Gujarati (WMT19) & \centdash & \centdash & 1,016 \\
          \hspace{10pt} Kazakh (WMT19) & \centdash & \centdash & 1,000 \\
          \hspace{10pt} Lithuanian (WMT19) & \centdash & \centdash & 1,000 \\
          \midrule
          \bf Document-level & & & \\
          \hspace{10pt} German (WMT08-19) & 366 & 69 & 145\\
          \hspace{10pt} Russian (WMT15-19) & 249 & 115 & 196 \\
          \hspace{10pt} Chinese (WMT17-19) & 123 & 222 & 163\\
         \bottomrule
    \end{tabular}
    }
    \caption{\label{tab:data}Number of sentences and documents per split for the languages used throughout this paper.}
\end{table}

\subsection{Translations}

We obtain the MT part of the data set by translating the non-English source texts to English by using Google Translate or DeepL. The translations were obtained in November-December 2022, except for the German translations, which we take from \newcite{van-der-werff-etal-2022-automatic} and were obtained in November 2021.\footnote{We translated the German test set in April 2023 with both Google and DeepL and compared them to the original translation of November 2021. We found BLEU scores of 98.27 and 98.54 for Google and DeepL, respectively, leading us to conclude that there are no substantial differences between the two versions of the MT systems. }

The data set we feed to our classification model is built by selecting exactly one human translation and one machine translation (either Google or DeepL) per source text. This way, we ensure there is no influence of the domain of the texts, while simultaneously ensuring a perfectly balanced data set for each experiment. Note that this also means that we actually train and test on twice as much data as is reported in Table~\ref{tab:data}. \textit{Target-only} or \emph{monolingual} classifiers are trained only on the English translations, while \textit{source + target} or \emph{multilingual} classifiers are trained on 
both
the source text and the English translation thereof. For evaluation, we also use MT outputs from selected WMT2019's submissions.\footnote{Details in Appendix~\ref{sec:appendix_wmt} (Table~\ref{tab:wmt_systems}).}

\begin{table*}[!t]
    \centering
    \resizebox{\textwidth}{!}{

    \input{table_versions/blue_red.tex}
    }
    \caption{\label{tab:diff_sl}Accuracies for the target-only \debertavt{} model when training on English translations (by Google or DeepL) from German and testing on translations from a different source language and different MT system on the test set. For German we report results both on the development (de-d) and test (de-t) sets. DeepL does not offer translations from Gujarati or Kazakh. }
\end{table*}

\subsection{Classifiers} We follow previous work \cite{bhardwaj-etal-2020-human,van-der-werff-etal-2022-automatic} in fine-tuning a pre-trained language model on our task. We use \debertavt{} \cite{debertav3} for the target-only classifiers since this was the best model by~\newcite{van-der-werff-etal-2022-automatic}.
For the source + target classifiers we test \mbert{} \cite{bert}, \mbart{} \cite{bart}, \xlmr{} \cite{xlm} and \mdeberta{} \cite{debertav3},
while \newcite{bhardwaj-etal-2020-human} only used \mbert{} and \xlmr{}.

We fine-tuned our pre-trained language models by using the Transformers library from HuggingFace \cite{wolf-etal-2020-transformers}. We use the \emph{ForSequenceClassification} implementations, both for the target-only as well as the source + target models. For the latter, this means that the source and target are concatenated by adding the \textsc{[SEP]} special character, which is the default implementation when providing two input sentences. We did experiment with adding source and target in the reverse order, but did not obtain improved performance. We did not experiment with adding a language tag to the source text.

\subsection{Experimental details} 
The results for Experiment 1 were obtained without any hyper-parameter tuning - we simply took the settings of \newcite{van-der-werff-etal-2022-automatic}. For finding the best multi-lingual language model (Experiment 2), we did perform a search over batch size and learning rate on the development set. We performed separate searches for the Google and DeepL translations, as well as the monolingual and bilingual settings. The final settings are shown in Table~\ref{tab:hyperparams} in Appendix~\ref{sec:appendix_hyperparam}.
For Experiment 3 and Experiment 4, we used the settings of the previously found best models. For the document-level systems in Experiment 5 we used the hyperparameters listed in Table~\ref{tab:doc_hyperparams} in Appendix~\ref{sec:appendix_hyperparam}. 
Reported accuracies are averaged over three runs for the sentence-level experiments (Exp 1--4) and over ten runs for the document-level experiments (Exp 5). Standard deviations (generally in range 0.2 - 2.0) are omitted for brevity, except for the document-level experiments, since they tend to be higher in the latter setting. All our code, data and results are publicly available.\footnote{\url{https://github.com/Malina03/macocu-ht-vs-mt/}}

\section{Results}

\subsection{Experiment 1: Testing on Translations from Different Source Languages}\label{sec:exp1}

In our first experiment (with results in Table~\ref{tab:diff_sl}), we analyse the performance of our classifier when testing a target-only model on English translations from a different source language. 
Here, the machine translations for training our classifier come from Google
or
DeepL, while we evaluate on translations from Google, DeepL and the two top-ranked (WMT1, WMT2) and two bottom-ranked (WMT3, WMT4) WMT2019 submissions \cite{barrault-etal-2019-findings}. See Appendix~\ref{sec:appendix_wmt} for additional details on these WMT submissions.

The results in Table~\ref{tab:diff_sl} show that human and machine translations from a different source language can still be 
reasonably well distinguished.
For certain languages, we are even very close to the performance on German (the original source language).
The other languages do seem to show an influence of the source language, as the accuracies are generally slightly lower, but are usually still comfortably above chance-level. However, there are a few cases were the classifier now performs \emph{below} chance level. This happened only for the bottom-ranked WMT systems (WMT3 and WMT4), which might not be representative of high-quality MT systems.

\begin{figure*}[!t]
\centering
\includegraphics[scale=0.6]{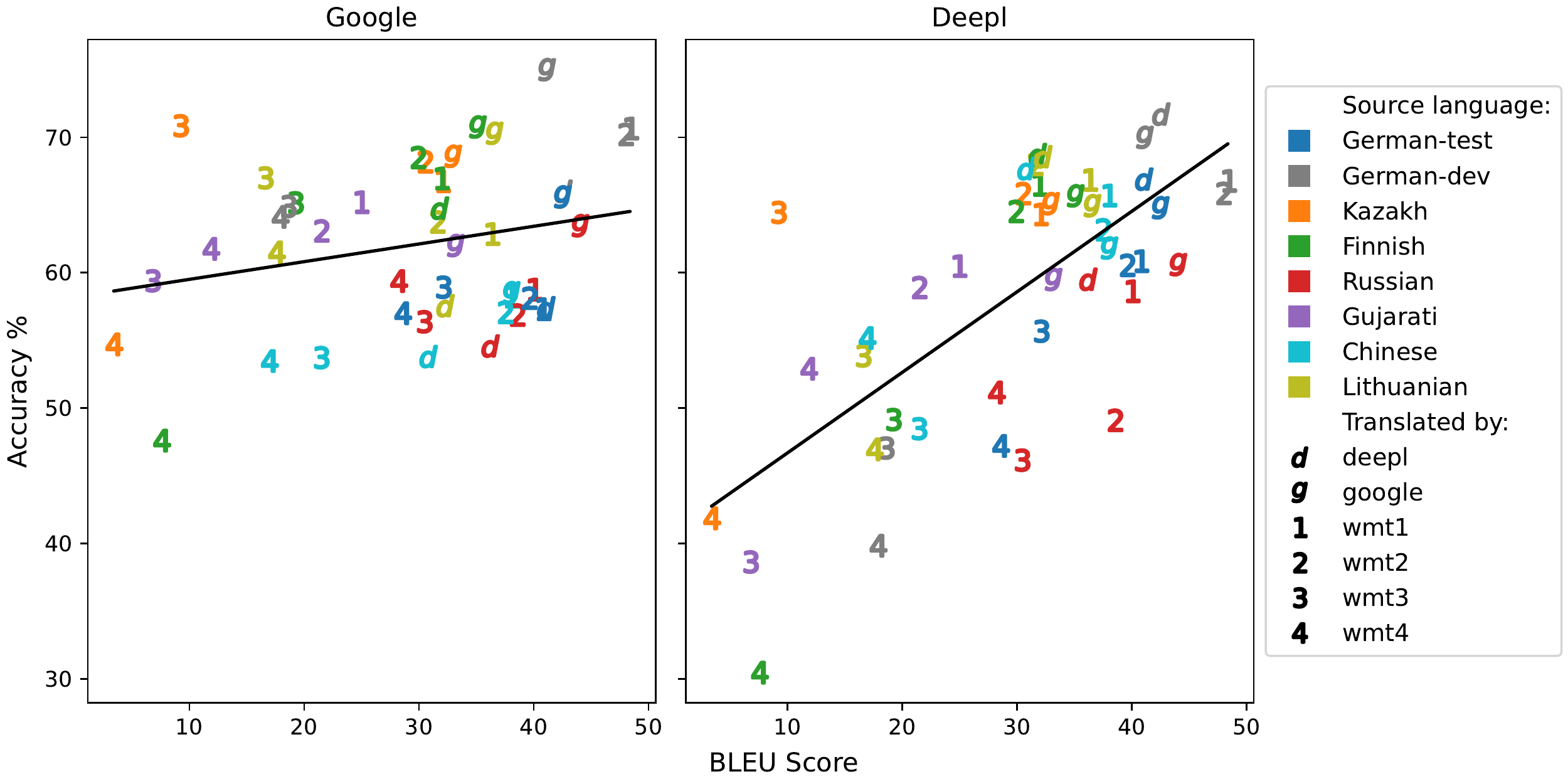}

\caption{\label{fig:bleu_google}Accuracy versus BLEU scores for each system in Table~\ref{tab:diff_sl}, using Google or DeepL translations during training.}
\end{figure*}

\paragraph {MT quality vs accuracy} We are also interested in how the quality of the translations influences accuracy scores. Since we have the human (reference) translations, we plotted the accuracy score 
of our classifier
versus an automatic MT evaluation metric, BLEU \cite{papineni-etal-2002-bleu}, in Figure~\ref{fig:bleu_google}.\footnote{Plots for COMET \cite{comet} instead of BLEU are in Appendix~\ref{sec:app_add}.} What is quite striking here is that we actually obtain an increased performance for higher-quality translations. When training on DeepL translations we actually find a significant correlation between accuracy and BLEU ($R=0.696$, $p<0.0001$), though for Google translations we did not ($R=0.249$, $p=0.094$). Intuitively, it should be easier to distinguish between low-quality MT and HT, so this is likely a side-effect of training on the high-quality 
translations from Google and DeepL. We consider this an important lesson for future work: if a classifier learns to distinguish high-quality MT from HT, this does not mean that distinguishing lower-quality MT comes for free.

\begin{table}[!b]
    \centering
    \setlength{\tabcolsep}{3pt}
    \resizebox{\columnwidth}{!}{
    \begin{tabular}{lcccc}
         \toprule
         & \multicolumn{2}{c}{\bf Google} & \multicolumn{2}{c}{\bf DeepL} \\
         & tgt-only & src + tgt  &  tgt-only & src + tgt   \\
         \midrule
         \debertavt & \bf 75.0 & \centdash & \bf 71.7 & \centdash \\ 
         \mbert & 65.9 & 71.7 & 65.5 & 66.1 \\ 
         \mbart & 69.3 & 71.7 & 61.9 & 68.1 \\ 
         \xlmr  & 66.0 & 69.3 & 62.4 & 66.9 \\ 
         \mdeberta & 70.4 & \bf 74.9 & 65.1 & \bf 71.8 \\ 
         \bottomrule
    \end{tabular}
    }
    \caption{\label{tab:best_multi}Development set accuracies of the best monolingual LM by~\newcite{van-der-werff-etal-2022-automatic} (\debertavt) and multi-lingual LMs, comparing the use of target-only and source + target data. The classifiers are trained and evaluated on the German--English data (Google or DeepL). Best result per column in bold.}

\end{table}

\subsection{Experiment 2: Source-only vs Source+Target Classifiers}
In our second experiment, we aim to determine whether having access to the source sentence improves classification performance. We test a variety of multilingual LMs, comparing their performance when having access only to the translation (target-only) to when also having access to the source sentence (source + target). Table~\ref{tab:best_multi} shows that accuracies indeed clearly improve for all of the tested LMs, with \mdeberta{} being the multilingual LM that leads to the highest accuracy. Note that this model performs similarly to the best target-only monolingual LM (\debertavt{}, with the scores taken from \newcite{van-der-werff-etal-2022-automatic}) on the development set, likely due to the higher quality of the latter LM for English. However, on the test set (also shown in Table~\ref{tab:cross}), which was never seen during development of the classifiers, the multilingual model is actually clearly superior (72.3\% versus 65.6\%).

\subsection{Experiment 3: Cross-system Evaluation} 

The study of \newcite{van-der-werff-etal-2022-automatic} showed that MT vs HT classifiers are sensitive to the MT system that was used to generate the training translations, as performance dropped considerably when doing a cross-system evaluation. However, we hypothesize that giving the classifier access to the source sentence will make it more robust to seeing translations from different MT systems at training and test times.

We show the results of the cross-MT system evaluation for the best performing target-only (\debertavt) and source + target (\mdeberta) models in Table~\ref{tab:cross}. For training on Google and testing on DeepL, we still see a considerable drop in performance for the source + target model (around 9 points in both the dev and test sets for both the target-only and source + target classifiers). However, when training on DeepL and testing on Google, we do see a clear effect on the test set: the target-only model dropped 2.1\% in accuracy (66.9 $\rightarrow$ 64.8), while the source + target model actually improved by 0.7\% (72.0 $\rightarrow$ 72.7).

\subsection{Experiment 4: Training on Multiple Source Languages}

Here, we investigate if we can actually combine training data from different source languages to improve performance. We run experiments for German, Russian and Chinese for both the target-only and the source + target model, of which the results are shown in Table~\ref{tab:multi_train}. Having additional training data from different source languages clearly helps, even for the multilingual source + target model. The only exception is the experiment on Chinese for the multilingual model, as the best performance (68\%) is obtained by only training on the Chinese training data.\footnote{The best performance on Chinese, in general, was, surprisingly, obtained by the target-only model (76.1\% accuracy).}
There does seem to be a diminishing effect of incorporating training data from different source languages
though, as the best score is only once obtained by combining all three languages as training data. Nevertheless, given the improved performance for even only small amounts of additional training data (Chinese has only 1,756 training instances), we see this as a promising direction for future work.

\begin{table}[!t]
    \centering
    \setlength{\tabcolsep}{3pt}
    \resizebox{\columnwidth}{!}{
    \begin{tabular}{l|cccc}
         \toprule
      \multicolumn{1}{c|}{\bf Evaluated on $\rightarrow$} & \multicolumn{2}{c}{\textbf{Dev}} & \multicolumn{2}{c}{\textbf{Test}} \\
        \textbf{$\downarrow$ Trained on} & \bf Google &  \bf DeepL & \bf Google &  \bf DeepL \\
        \midrule
         \debertavt &  & & &   \\
        \hspace{12pt} Google & \bf 75.0  & 66.0 & \bf 65.6 & 57.4   \\
       \hspace{12pt} DeepL & 70.0  & \bf 71.7 & 64.8 & \bf 66.9  \\
        \midrule
        \mdeberta &  & & &   \\
        \hspace{12pt}Google & \bf 74.9  & 66.2 & 72.3 & 63.8   \\
        \hspace{12pt}DeepL & 71.3  & \bf 71.8 & \bf 72.7 & \bf 72.0   \\
         \bottomrule
    \end{tabular}
    }
    \caption{\label{tab:cross}Dev and test set accuracies of \debertavt{} (target-only) and \mdeberta{} (source + target) when trained and evaluated on Google and DeepL. First two rows of results taken from \newcite{van-der-werff-etal-2022-automatic}. Best score per column and classifier in bold.}
\end{table}

\begin{table}[!t]
    \centering
    \setlength{\tabcolsep}{4pt}
    \resizebox{\columnwidth}{!}{
    \begin{tabular}{lcccccc}
      \toprule
      \multicolumn{1}{r}{\bf Eval $\rightarrow$} & \multicolumn{3}{c}{\textbf{\debertavt}} & \multicolumn{3}{c}{\textbf{\mdeberta}} \\
        \textbf{$\downarrow$ Train} & \bf de & \bf zh & \bf ru & \bf de & \bf zh & \bf ru \\
         \midrule
         German (de)  & 65.6 & 64.2 & 63.3 & 72.3 & 55.1 & 66.1 \\
         Chinese (zh) & 58.1 & 75.4 & 53.4 & 63.5 & \bf 68.0 & 61.6 \\
         Russian (ru) & 56.7 & 52.3 & 63.1 & 64.3 & 56.7 & 69.0 \\
         \midrule
         de + zh      & \bf 66.6 & \bf 76.1 & 63.7 & 72.7 & 66.2 & 68.7 \\
         de + ru      & 66.3 & 62.0 & 67.1 & \bf 73.6 & 58.5 & \bf 71.6 \\
         ru + zh      & 59.7 & 75.5 & 66.2 & 66.3 & 66.0 & 69.3 \\
         de + zh + ru & 66.5 & 75.2 & \bf 68.1 & 72.8 & 65.8 & 71.3 \\
         
         \bottomrule
    \end{tabular}
    }
    \caption{\label{tab:multi_train}Test set accuracies on discriminating between HT and Google Translate with \debertavt{} (target-only) and \mdeberta{} (source + target) when training on data from one versus multiple source languages. Best score per column shown in bold.}
\end{table}

\subsection{Experiment 5: Sentence- vs Document-level}
We perform a similar experiment as \newcite{van-der-werff-etal-2022-automatic} by testing our classifiers on the document-level, as the WMT data sets include this information. We expect that the task is (a lot) easier if the classifier has access to full documents instead of just sentences. We test this with both the best monolingual (\debertavt) and  multi-lingual (\mdeberta) models on Google translations from German. 

\paragraph{Truncation} DeBERTa models can in principle work with sequence lengths up to 24,528 tokens, but that does not mean this is optimal, especially when taking speed and memory requirements into account. In Table~\ref{tab:truncation} we compare accuracies for different values of maximum length, or in other words, different levels of truncation. For \debertavt{}, the preferred truncation value is 1,024 tokens, while for \mdeberta{} we opt for 3,072. For both models, the input documents are barely truncated. The larger value for \mdeberta{} is expected, as those experiments have roughly twice the amount of input tokens (source- + target-language data versus just target data). Lengths of 3,072 (\debertavt) or 4,096 (\mdeberta) did not fit into our GPU memory (NVIDIA V100) even with a batch size of 1, but looking at the scores and truncation percentages, this does not seem to be an issue.

\begin{table}[!t]
    \centering
    \setlength{\tabcolsep}{5pt}
    \resizebox{\columnwidth}{!}{
    \begin{tabular}{r|rrr|rrr}
         \toprule
        \bf max & \multicolumn{3}{c|}{\bf \debertavt} & \multicolumn{3}{c}{\bf \mdeberta} \\
        \bf length & \bf T (\%) & \bf T (avg) & \bf Acc. & \bf T (\%) & \bf T (avg) & \bf Acc. \\
        \midrule
        512 & 38 & 132 & 79.4 & 77 & 793 & 75.9\\
        768 & 17 & 62 & 95.0 & 62 & 617 & 80.1\\
        1,024 & 8 & 32 & \bf 96.4 & 50 & 472 & 85.3\\
        2,048 & 0.8 & 4 & 93.4 & 16 & 155 & 89.7\\
        3,072 & 0.0 & 0.0 & \multicolumn{1}{c|}{---} & 5 & 20 & \bf 91.9 \\
         \bottomrule
    \end{tabular}
    }
    \caption{\label{tab:truncation}Document-level accuracies (\emph{Acc.}) for different values of maximum length (number of tokens) on the German development set, trained on German data. \emph{T (\%)} indicates the percentage of training documents that were truncated. \emph{T (avg)} indicates the average amount of tokens that were truncated across the training set. Best score per classifier in bold.}
\end{table}

\paragraph{Evaluation} We evaluate the models using the preferred truncation settings found above.\footnote{Hyperparameters used are shown in Appendix~\ref{sec:appendix_hyperparam} (Table~\ref{tab:doc_hyperparams}).} We train on either just German, or German, Russian, and Chinese data, and evaluate on the German data.\footnote{Results for Russian and Chinese are in Appendix~\ref{sec:app_add} (Table~\ref{tab:document_ru_zh}).} We evaluate the performance on three different classifiers: (i) applying the best sentence-level model on the documents sentence by sentence, and taking the majority vote, (ii) simply training on the documents instead of sentences and (iii) fine-tuning the best sentence-level model on documents.
The latest classifier is motivated by the fact that there are much fewer document-level training instances than there are of sentence-level (Table~\ref{tab:data}).

\paragraph{Document-level classifiers} The results are shown in Table~\ref{tab:document}, which allows us to draw the following conclusions. For one, fine-tuning the sentence-level model on documents is clearly preferable over simply training on documents, while also comfortably outperforming the majority vote baseline. 
Fine-tuning not only leads to the highest accuracies, but also to the lowest standard deviations, indicating that this classifier is more stable than the other two.
Second, we confirm our two previous findings: the models can improve performance when training on texts from a different source language (Chinese and Russian in this case) and the models clearly benefit from having access to the source text itself during training and evaluation.

\section{Conclusion}

This paper has investigated the discrimination between neural machine translation (NMT) and human translation (HT) in multilingual scenarios, using as classifiers monolingual and multilingual language models (LMs) that are fine-tuned with small amounts of task-specific labelled data. 

We have found out that a monolingual classifier trained on English translations from a given source language still performs well above chance on English translations from other source languages.
Using a multilingual LM and therefore having access also to the source sentence results overall in better performance than an equivalent LM that only has access to the target sentence. Such a classifier seems more robust in a cross-system situation, i.e. when the MT systems used to train and evaluate the classifier are different.
Moreover, as task-specific data is limited, we experimented with (i) training on data from different source languages and (ii) training on the document-level instead of the sentence-level, with improved performance in both settings.

\begin{table}[!t]
    \centering
    \setlength{\tabcolsep}{2.5pt}
    \resizebox{\columnwidth}{!}{
    \begin{tabular}{lcccc}
         \toprule
         \multicolumn{1}{c}{\bf Trained on $\rightarrow$}  & \multicolumn{2}{c}{\bf German (de) } & \multicolumn{2}{c}{\bf de + ru + zh} \\
        & \textbf{\deb} & \textbf{\mdeb} & \textbf{\deb} & \textbf{\mdeb} \\
        \midrule
        Majority vote & 68.5 \stddev{8.7} & 73.1 \stddev{5.7}& 75.6 \stddev{6.7} & 76.5 \stddev{4.7}\\
        Doc-level & 62.6 \stddev{3.6} & 75.3 \stddev{3.9} & 67.3 \stddev{10.7} & 83.0 \stddev{2.2} \\
        Doc-level (ft) & \textbf{81.1} \stddev{2.7} & \textbf{86.0} \stddev{1.2} & \textbf{87.0} \stddev{2.6} & \textbf{88.7} \stddev{1.4} \\
         \bottomrule
    \end{tabular}
    }
    \caption{\label{tab:document}Document-level accuracies and standard deviations with \debertavt{} (target-only, denoted as \deb{}) and \mdeberta{} (source + target, \mdeb{}) when evaluating on the test that has German as the source language using Google as the MT system. Best result per column shown in bold.}
\end{table}

\subsection{Future work} In this work, we took an important step toward developing an accurate, reliable, and accessible classifier that can distinguish between HT and MT. There are, of course, still many research directions to explore, in particular regarding combining different source languages and MT systems during training. Moreover, in many practical applications, it is unknown whether a text is actually a translation, as it can also be an original text. Therefore, in future work, we aim to develop a classifier that can distinguish between original texts, human translations, and machine translations.

\section*{Acknowledgements}

The authors received funding from the European Union's Connecting Europe Facility 2014-2020 - CEF Telecom, under Grant Agreement No. INEA/CEF/ICT/A2020/2278341 (MaCoCu). This communication reflects only the authors' views. The Agency is not responsible for any use that may be made of the information it contains. We thank the Center for Information Technology of the University of Groningen for providing access to the Peregrine high performance computing cluster. Finally, we thank all our MaCoCu colleagues for their valuable feedback throughout the project.

\bibliography{anthology,custom}
\bibliographystyle{eamt23}

\appendix

\section{WMT MT Systems\label{sec:appendix_wmt}}

Table~\ref{tab:wmt_systems} shows the specific WMT19 systems that were used during Experiment 1. \newcite{barrault-etal-2019-findings} did not specify which specific online systems were used.

\section{Hyperparameters\label{sec:appendix_hyperparam}}

Sentence-level hyperparameters used in our experiments are shown in Table~\ref{tab:hyperparams}, while the document-level settings are shown in Table~\ref{tab:doc_hyperparams}.

\vspace{1cm}

\begin{table}[!htbp]
 
    \begin{minipage}{\textwidth}
       \centering
    \begin{tabular}{lcccc}
         \toprule
         & \bf WMT1 & \bf WMT2 & \bf WMT3 & \bf WMT4 \\
          \midrule                  
de & \newcite{ng-etal-2019-facebook}   & \newcite{rosendahl-etal-2019-rwth}    & \newcite{molchanov-2019-promt}            & online-X                        \\
fi & \newcite{xia-etal-2019-microsoft} & online-Y                            & \newcite{bicici-2019-machine}             & \newcite{pirinen-2019-apertium}   \\
 gu & \newcite{li-etal-2019-niutrans}   & \newcite{bawden-etal-2019-university} & \newcite{goyal-sharma-2019-iiit}          & \newcite{mondal-etal-2019-ju}     \\
 kk & online-B                        & \newcite{li-etal-2019-niutrans}       & \newcite{briakou-carpuat-2019-university} & \newcite{budiwati-etal-2019-dbms} \\
 lt & \newcite{bei-etal-2019-gtcom}     & \newcite{pinnis-etal-2019-tildes}     & JUMT                                    & online-X                        \\
 ru & \newcite{ng-etal-2019-facebook}   & online-G                            & online-X                                & \newcite{dabre-etal-2019-nicts}   \\
 zh & \newcite{sun-etal-2019-baidu}     & \newcite{guo-etal-2019-kingsofts}     & \newcite{li-specia-2019-comparison}       & online-X   \\
          \bottomrule
    \end{tabular}
    
    \caption{\label{tab:wmt_systems}WMT systems used in our analysis. WMT1 and WMT2 are the two top-ranked systems, while WMT3 and WMT4 are the two bottom-ranked systems. The JUMT system did not submit a paper.}
    \end{minipage}
\end{table}

\begin{table}[!hbtp]

\begin{minipage}{\textwidth}
\centering
\begin{tabular}{lcccccccc}
\toprule
 & \multicolumn{4}{c}{\bf Monolingual} & \multicolumn{4}{c}{\bf Multilingual} \\
 & \multicolumn{2}{c}{\bf Learning Rate} & \multicolumn{2}{c}{\bf Batch Size} & \multicolumn{2}{c}{\bf Learning Rate} & \multicolumn{2}{c}{\bf Batch Size} \\
 & \bf Google & \bf DeepL & \bf Google & \bf DeepL & \bf Google & \bf DeepL & \bf Google & \bf DeepL \\
 \midrule
\debertavt & \expnot{1}{-5} & \expnot{1}{-5} & 32 & 32 &  \centdash  &  \centdash  &  \centdash  &  \centdash  \\
\mbert & \expnot{1}{-5} & \expnot{1}{-5} & 16 & 32 & \expnot{1}{-5} & \expnot{1}{-5} & 16 & 32 \\
\mbart & \expnot{1}{-5} & \expnot{5}{-6} & 16 & 32 & \expnot{5}{-6} & \expnot{1}{-5} & 16 & 16 \\
\xlmr & \expnot{1}{-5} & \expnot{1}{-5} & 16 & 32 & \expnot{1}{-5} & \expnot{1}{-5} & 16 & 16 \\
\mdeberta & \expnot{1}{-5} & \expnot{1}{-5} & 32 & 32 & \expnot{5}{-5} & \expnot{1}{-5} & 32 & 16 \\
\bottomrule
\end{tabular}
    \caption{\label{tab:hyperparams}Final hyper-parameter settings for the models used throughout the paper. We experimented with a batch size of $\{16, 32, 64\}$ and a learning rate of $\{$\expnot{1}{-4}, \expnot{1}{-5}, \expnot{5}{-5}, \expnot{1}{-6}, \expnot{5}{-6}$\}$.}
\end{minipage}
\end{table}

\begin{table*}
\centering
\begin{tabular}{lcccc}
\toprule
 & \bf Max Sequence Length & \bf Learning Rate & \bf Batch Size & \bf Gradient Accumulation \\
 \midrule
\debertavt & 1,024 & \expnot{1}{-5} & 2 & 8   \\
\mdeberta & 3,072 & \expnot{1}{-5} & 1 & 8 \\
\bottomrule
\end{tabular}
    \caption{\label{tab:doc_hyperparams}Final hyper-parameter settings for the models trained on the document level. We experimented with batch sizes of $\{1,2,4,8\}$ and different gradient accumulation values such that the effective batch size was at most $16$ due to the hardware limitations. The learning rates tested were $\{$\expnot{1}{-6}, \expnot{2}{-6}, \expnot{5}{-6}, \expnot{1}{-5} $\}$.}
\end{table*}

\begin{table*}
    \centering
     \resizebox{\textwidth}{!}{
    \begin{tabular}{lcccccccc}
         \toprule
         \multicolumn{1}{c}{\bf Tested on $\rightarrow$} & \multicolumn{4}{c}{\bf Russian} & \multicolumn{4}{c}{\bf Chinese} \\
         \midrule
         \multicolumn{1}{c}{\bf Trained on $\rightarrow$}  & \multicolumn{2}{c}{\bf German (de) } & \multicolumn{2}{c}{\bf de + ru + zh} & \multicolumn{2}{c}{\bf German (de) } & \multicolumn{2}{c}{\bf de + ru + zh} \\
         & \textbf{\deb} & \textbf{\mdeb} & \textbf{\deb} & \textbf{\mdeb} & \textbf{\deb} & \textbf{\mdeb} & \textbf{\deb} & \textbf{\mdeb} \\
        \midrule
        Majority vote & 67.2 \stddev{9.1} & 68.8 \stddev{2.2} & 71.0 \stddev{5.5} & 69.5 \stddev{4.7} & 62.5 \stddev{10.8} & 53.8 \stddev{5.2} & 92.6 \stddev{11.9} & 65.2 \stddev{4.6}\\
        Doc-level & 58.7 \stddev{3.8} & \textbf{73.3} \stddev{2.7} & 66.0 \stddev{6.4} & 71.7 \stddev{2.4} & 53.6 \stddev{3.6} & \textbf{69.0} \stddev{8.8} & 84.8 \stddev{13.6} & \textbf{84.4} \stddev{6.9}\\
        Doc-level (ft) & \textbf{78.4} \stddev{1.8} & 72.8 \stddev{2.4} & \textbf{75.3} \stddev{2.3} & \textbf{80.6} \stddev{1.5} & \textbf{76.5} \stddev{6.4} & 52.6 \stddev{1.2} & \textbf{96.0} \stddev{0.9} & 78.0 \stddev{1.8}\\
         \bottomrule
    \end{tabular}
     }
    \caption{\label{tab:document_ru_zh}Document-level accuracies when evaluating on test sets where the source language is either Russian or Chinese (see Table~\ref{tab:document} for results on German). We train source-only \debertavt{} (\deb) and source + target \mdeberta{} (\mdeb) models on either German, or German, Russian and Chinese combined. }
\end{table*}

\section{Additional Results\label{sec:app_add}}

\begin{figure*}[!htb]
\centering
\includegraphics[scale=0.6]{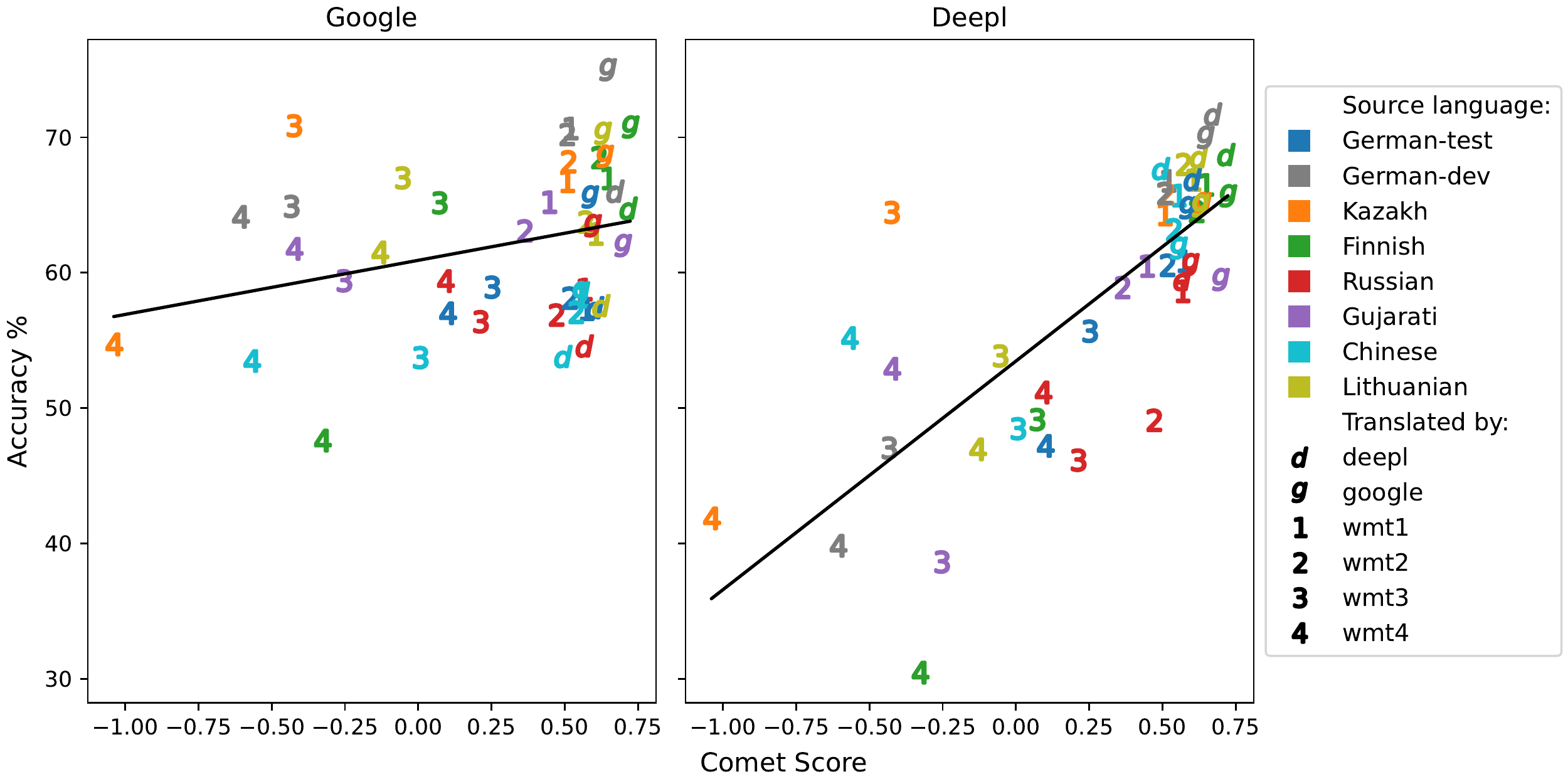}
\caption{\label{fig:comet_google_deepl}Accuracy versus COMET scores for each system in Table~\ref{tab:diff_sl}, using Google (left) or DeepL (right) translations during training. Accuracy versus BLEU scores can be found in Figure~\ref{fig:bleu_google}.}
\end{figure*}

Figure~\ref{fig:comet_google_deepl} shows
additional results for Experiment 1, specifically scatter plots of the accuracy of the classifier versus COMET scores for each system for both Google and DeepL.
This complements Figure~\ref{fig:bleu_google} in Section~\ref{sec:exp1}, in which BLEU was used instead of COMET. The trends are very similar in both figures.

Table~\ref{tab:document_ru_zh} shows additional evaluation results on document-level classification (Experiment 5), as opposed to Table~\ref{tab:document} in which we evaluated on the test set that has German as the source language. We observe that fine-tuning the sentence-level model is still generally preferable, though there are few cases in which just training on documents resulted in the best performance. A curious observation is that for Chinese including the source text does generally not lead to improved performance, while this is not the case for Russian and German.

\end{document}

%% file: table_versions/blue_red.tex
\begin{tabular}{l|cccccccc|cccccccc}
     \toprule
      & \multicolumn{8}{c|}{\textbf{Trained on Google translations}} & \multicolumn{8}{c}{\textbf{Trained on DeepL translations}} \\
      \midrule
     \textbf{$\downarrow$ Eval} & \textbf{de-d} & \textbf{de-t} & \textbf{fi} & \textbf{gu} & \textbf{kk} & \textbf{lt} & \textbf{ru} & \textbf{zh} & \textbf{de-d} & \textbf{de-t} & \textbf{fi} & \textbf{gu} & \textbf{kk} & \textbf{lt} & \textbf{ru} & \textbf{zh}  \\
     \midrule

        DeepL  & \cellcolor{blue!79.78!red} \color{white}{ 66.0 } & \cellcolor{blue!60.45!red} \color{white}{ 57.4 } & \cellcolor{blue!77.08!red} \color{white}{ 64.8 } &  \centdash  &  \centdash  & \cellcolor{blue!60.9!red} \color{white}{ 57.6 } & \cellcolor{blue!54.16!red} \color{white}{ 54.6 } & \cellcolor{blue!52.36!red} \color{white}{ 53.8 } & \cellcolor{blue!92.58!red} \color{white}{ 71.7 } & \cellcolor{blue!81.8!red} \color{white}{ 66.9 } & \cellcolor{blue!85.84!red} \color{white}{ 68.7  } &  \centdash  &  \centdash  & \cellcolor{blue!85.62!red} \color{white}{ 68.6 } & \cellcolor{blue!65.17!red} \color{white}{ 59.5  } & \cellcolor{blue!83.6!red} \color{white}{  67.7} \\
        Google  & \cellcolor{blue!100.0!red} \color{white}{ 75.0 } & \cellcolor{blue!78.88!red} \color{white}{ 65.6 } & \cellcolor{blue!90.56!red} \color{white}{ 70.8 } & \cellcolor{blue!70.79!red} \color{white}{ 62.0 } & \cellcolor{blue!85.62!red} \color{white}{ 68.6 } & \cellcolor{blue!89.44!red} \color{white}{ 70.3 } & \cellcolor{blue!74.16!red} \color{white}{ 63.5 } & \cellcolor{blue!62.92!red} \color{white}{ 58.5  } & \cellcolor{blue!88.76!red} \color{white}{ 70.0 } & \cellcolor{blue!77.08!red} \color{white}{   64.8 } & \cellcolor{blue!79.1!red} \color{white}{ 65.7 } & \cellcolor{blue!65.17!red} \color{white}{ 59.5 } & \cellcolor{blue!77.75!red} \color{white}{ 65.1 } & \cellcolor{blue!77.53!red} \color{white}{ 65.0 } & \cellcolor{blue!67.64!red} \color{white}{ 60.6 } & \cellcolor{blue!70.34!red} \color{white}{ 61.8 } \\
        WMT\textsubscript{1}  & \cellcolor{blue!60.22!red} \color{white}{ 57.3 } & \cellcolor{blue!90.34!red} \color{white}{ 70.7 } & \cellcolor{blue!82.02!red} \color{white}{ 67.0 } & \cellcolor{blue!77.98!red} \color{white}{ 65.2 } & \cellcolor{blue!81.57!red} \color{white}{ 66.8 } & \cellcolor{blue!72.81!red} \color{white}{ 62.9 } & \cellcolor{blue!63.6!red} \color{white}{ 58.8 } & \cellcolor{blue!62.25!red} \color{white}{ 58.2 } & \cellcolor{blue!68.31!red} \color{white}{ 60.9  } & \cellcolor{blue!81.57!red} \color{white}{  66.8 } & \cellcolor{blue!80.9!red} \color{white}{ 66.5 } & \cellcolor{blue!67.42!red} \color{white}{ 60.5 } & \cellcolor{blue!75.96!red} \color{white}{ 64.3 } & \cellcolor{blue!81.8!red} \color{white}{ 66.9 } & \cellcolor{blue!63.15!red} \color{white}{ 58.6 } & \cellcolor{blue!79.1!red} \color{white}{ 65.7 } \\
        WMT\textsubscript{2}  & \cellcolor{blue!62.02!red} \color{white}{ 58.1 } & \cellcolor{blue!89.21!red} \color{white}{ 70.2 } & \cellcolor{blue!85.39!red} \color{white}{ 68.5 } & \cellcolor{blue!73.26!red} \color{white}{ 63.1 } & \cellcolor{blue!84.72!red} \color{white}{ 68.2 } & \cellcolor{blue!74.83!red} \color{white}{ 63.8 } & \cellcolor{blue!59.33!red} \color{white}{ 56.9 } & \cellcolor{blue!59.78!red} \color{white}{ 57.1 } & \cellcolor{blue!67.64!red} \color{white}{ 60.6 } & \cellcolor{blue!79.55!red} \color{white}{  65.9 } & \cellcolor{blue!76.4!red} \color{white}{ 64.5 } & \cellcolor{blue!63.82!red} \color{white}{ 58.9 } & \cellcolor{blue!79.55!red} \color{white}{ 65.9 } & \cellcolor{blue!84.27!red} \color{white}{ 68.0 } & \cellcolor{blue!41.8!red} \color{white}{ 49.1 } & \cellcolor{blue!73.48!red} \color{white}{ 63.2 } \\
        WMT\textsubscript{3}  & \cellcolor{blue!63.82!red} \color{white}{ 58.9 } & \cellcolor{blue!77.3!red} \color{white}{ 64.9 } & \cellcolor{blue!77.98!red} \color{white}{ 65.2 } & \cellcolor{blue!64.94!red} \color{white}{ 59.4 } & \cellcolor{blue!90.79!red} \color{white}{ 70.9 } & \cellcolor{blue!82.02!red} \color{white}{ 67.0 } & \cellcolor{blue!58.2!red} \color{white}{ 56.4 } & \cellcolor{blue!52.13!red} \color{white}{ 53.7 } & \cellcolor{blue!56.63!red} \color{white}{ 55.7 } & \cellcolor{blue!37.3!red} \color{white}{  47.1 } & \cellcolor{blue!42.02!red} \color{white}{ 49.2 } & \cellcolor{blue!18.2!red} \color{white}{ 38.6 } & \cellcolor{blue!76.4!red} \color{white}{ 64.5 } & \cellcolor{blue!52.58!red} \color{white}{ 53.9 } & \cellcolor{blue!35.28!red} \color{white}{ 46.2 } & \cellcolor{blue!40.45!red} \color{white}{ 48.5 } \\
        WMT\textsubscript{4}  & \cellcolor{blue!59.55!red} \color{white}{ 57.0 } & \cellcolor{blue!75.51!red} \color{white}{ 64.1 } & \cellcolor{blue!38.43!red} \color{white}{ 47.6 } & \cellcolor{blue!70.34!red} \color{white}{ 61.8 } & \cellcolor{blue!54.38!red} \color{white}{ 54.7 } & \cellcolor{blue!69.66!red} \color{white}{ 61.5 } & \cellcolor{blue!64.94!red} \color{white}{ 59.4 } & \cellcolor{blue!51.69!red} \color{white}{ 53.5 } & \cellcolor{blue!37.53!red} \color{white}{ 47.2 } & \cellcolor{blue!20.9!red} \color{white}{  39.8 } & \cellcolor{blue!0.0!red} \color{white}{ 30.5 } & \cellcolor{blue!50.34!red} \color{white}{ 52.9 } & \cellcolor{blue!25.62!red} \color{white}{ 41.9 } & \cellcolor{blue!37.08!red} \color{white}{ 47.0 } & \cellcolor{blue!46.52!red} \color{white}{ 51.2 } & \cellcolor{blue!55.51!red} \color{white}{ 55.2 } \\
    \bottomrule
\end{tabular}